\documentclass[runningheads]{llncs}

\usepackage{eccv}

\usepackage{eccvabbrv}

\usepackage{graphicx}
\usepackage{amsmath}
\usepackage{amssymb}
\usepackage{booktabs}
\usepackage{multirow}
\usepackage{array}
\usepackage{tabularx}
\usepackage{cuted}
\usepackage{svg}
\usepackage{enumitem}

\usepackage{dsfont}
\usepackage{etoolbox}
\usepackage{color}

\newif\ifshowedits

\newcommand{\addeditor}[3]{%
  \definecolor{#1color}{rgb}{#3}
  \expandafter\newcommand\csname #1\endcsname[1]{%
  \ifshowedits
    {\color{#1color} ##1}%
  \else
    {##1}%
  \fi
  }%
  \expandafter\newcommand\csname #1rmk\endcsname[1]{%
  \ifshowedits
    {\color{#1color} {\bf [#2: ##1]}}
  \fi
  }%
  \expandafter\newcommand\csname #1rpl\endcsname[2]{%
  \ifshowedits
    {\color{#1color} ##1 \sout{##2}}
  \else
    {##1}
  \fi
  }%
}

\newcommand{\createtextvar}[1]{
  \expandafter\newcommand\csname #1\endcsname{%
  {\text{#1}}
}%
}
\newcommand{\PreserveBackslash}[1]{\let\temp=\\#1\let\\=\temp}
\newcolumntype{C}[1]{>{\PreserveBackslash\centering}p{#1}}
\newcolumntype{R}[1]{>{\PreserveBackslash\raggedleft}p{#1}}
\newcolumntype{L}[1]{>{\PreserveBackslash\raggedright}p{#1}}

\newcommand{\moretextwithfigures}{
\renewcommand{\topfraction}{1}
\renewcommand{\dbltopfraction}{1}
\renewcommand{\bottomfraction}{1}
\renewcommand{\textfraction}{.0}
\renewcommand{\floatpagefraction}{1}
\renewcommand{\dblfloatpagefraction}{1}
}

\newcommand{\mycomment}[1]{}

\newcommand{\calL}{{\cal L}}
\newcommand{\calM}{{\cal M}}
\newcommand{\calN}{{\cal N}}

\newcommand{\calR}{{\cal R}}

\newcommand{\calW}{{\cal W}}

\newcommand{\bg}{{\bf g}}

\newcommand{\bn}{{\bf n}}

\newcommand{\bv}{{\bf v}}
\newcommand{\bw}{{\bf w}}

\newcommand{\bz}{{\bf z}}

\newcommand{\vcomment}[1]{}

\addeditor{vincent}{VL}{0.0, 0.5, 0.0}
\addeditor{alexandre}{AC}{0.0, 0.0, 0.8}
\addeditor{nikos}{NP}{0.5, 0.0, 0.0}
\showeditsfalse
\moretextwithfigures

\usepackage[accsupp]{axessibility}  

\usepackage[pagebackref,breaklinks,colorlinks,citecolor=eccvblue]{hyperref}

\usepackage{orcidlink}

\usepackage[capitalize]{cleveref}
\crefname{section}{Sec.}{Secs.}
\Crefname{section}{Section}{Sections}
\Crefname{table}{Table}{Tables}
\crefname{table}{Tab.}{Tabs.}

\begin{document}


\title{GuidedRec: Guiding Ill-Posed Unsupervised Volumetric Recovery}

\author{Alexandre Cafaro\inst{1,2,3} \and 
Amaury Leroy\inst{1,2,3}\and 
Guillaume Beldjoudi\inst{3}\and
Pauline Maury\inst{2}\and
Charlotte Robert\inst{2}\and
Eric Deutsch\inst{2}\and
Vincent Grégoire\inst{3}\and
Vincent Lepetit\inst{4}\and
Nikos Paragios\inst{1}
}
\index{Cafaro, Alexandre}
\index{Leroy, Amaury}
\index{Beldjoudi, Guillaume}
\index{Maury, Pauline}
\index{Robert, Charlotte}
\index{Deutsch, Eric}
\index{Grégoire, Vincent}
\index{Lepetit, Vincent}
\index{Paragios, Nikos}

\authorrunning{A. Cafaro et al.}
%
\institute{TheraPanacea, Paris, France \and 
Gustave Roussy, Inserm 1030, Paris-Saclay University, Villejuif, France \and
Department of Radiation Oncology, Centre Léon Bérard, Lyon, France \and 
LIGM, Ecole des Ponts, Univ Gustave Eiffel, CNRS, France}

\maketitle


\begin{abstract}
We introduce a novel unsupervised approach to reconstructing a 3D volume from only two planar projections that exploits a previous\-ly-captured 3D volume of the patient. Such volume is readily available in many important medical procedures and previous methods already used such a volume. Earlier methods that work by deforming this volume to match the projections typically fail when the number of projections is very low as the alignment becomes underconstrained. We show how to use a generative model of the volume structures to constrain the deformation and obtain a correct estimate. Moreover, our method is not bounded to a specific sensor calibration and can be applied to new calibrations without retraining. We evaluate our approach on a challenging dataset and show it outperforms state-of-the-art methods. As a result, our method could be used in treatment scenarios such as surgery and radiotherapy while drastically reducing patient radiation exposure. 
\end{abstract}


\section{Introduction}
\label{sec:intro}

In medical treatments, the reconstruction of a partial 3D volume of the patient's body is essential for both initial treatment planning and ongoing monitoring of the patient's progress. This process, crucial for accurately targeting areas affected by tumors in interventions like radiotherapy or surgery, traditionally requires numerous X-ray projections, leading to significant radiation exposure and extended time commitments. In practice, however, only two projections are captured to reduce radiation exposure; they are used for rigid registration, a basic form of alignment. 

However, two projections currently fall short of providing the full details of 3D reconstructions. Despite attempts to accurately reconstruct anatomy with fewer projections, existing methods with just two projections lack the necessary detail for radiotherapy. We aim to develop a method that delivers precise reconstructions with only two projections, balancing the need for accuracy in treatment with reduced radiation exposure and shorter procedure times.

\begin{figure}[t]
\centering
\includegraphics[width=0.7\linewidth]{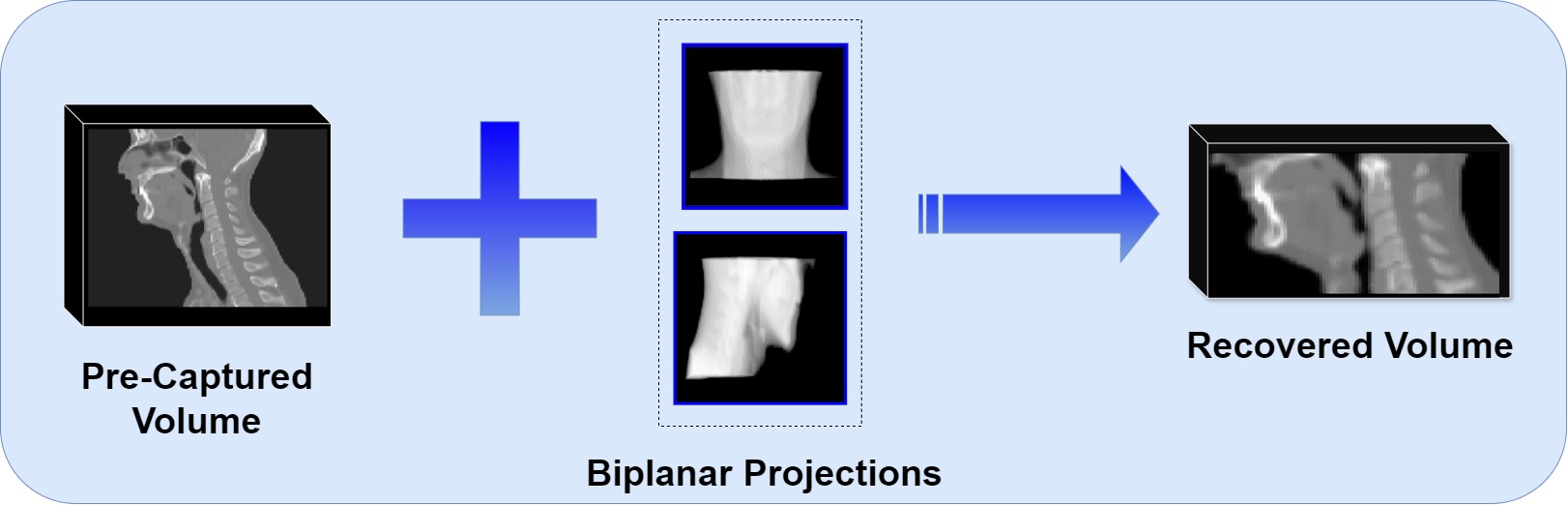}
\caption{The goal of our method is to recover an accurate 3D volume given two projections of the patient and a volume acquired at the beginning of the therapy. As discussed in the introduction, this ability  unlocks better therapy procedures. Combining correctly these sources of information is however challenging. \alexandrermk{to reference somewhere} \alexandrermk{too long i cant make it fit}\vincentrmk{why not the 2 projections between the + and the arrow?}
}
\label{fig:teaser}
\end{figure}

\begin{figure*}[t]
\centering
\includegraphics[width=\linewidth]{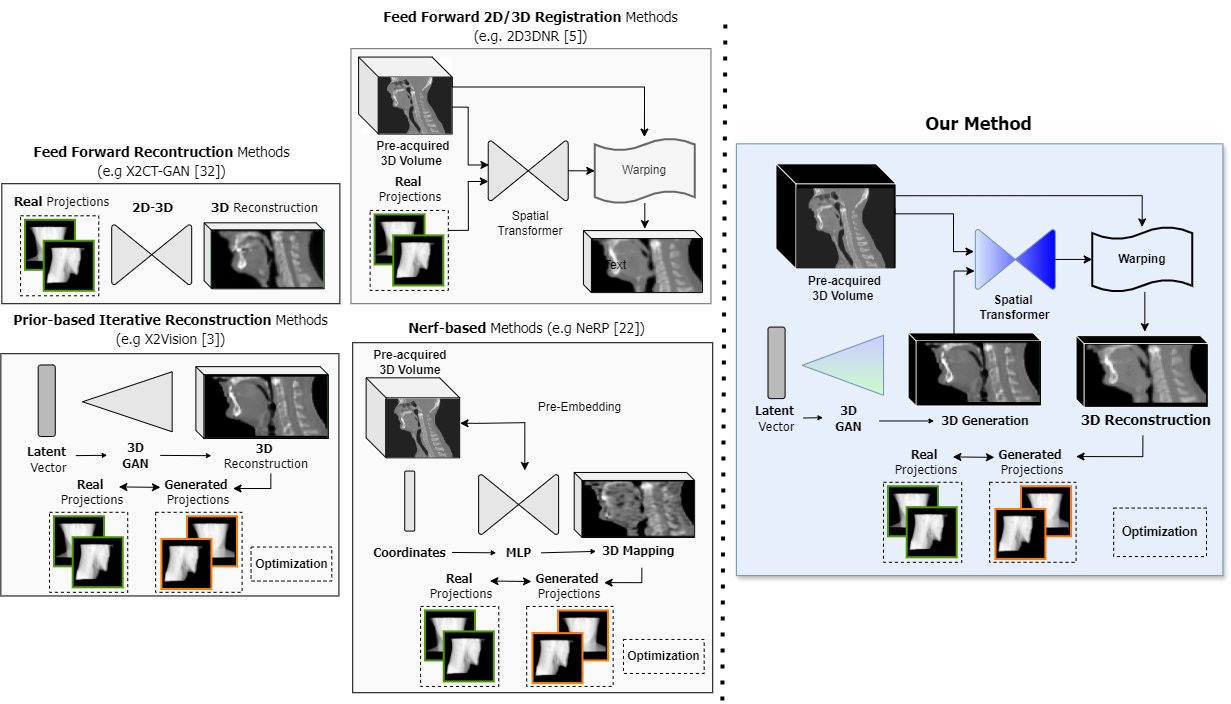} 
\caption{\textbf{Current methods vs our method.} 
\textbf{(1)} Feed-forward reconstruction~\cite{henzler2018single,shen2019patient,ying2019x2ct, jiang2022mfct, tan2022xctnet, Lu_2022, tan2023semi, zhang2023xtransct,  shen2022geometry, wang2023tpg}  
directly predict a volume from a set of projections. Their performance significantly degrades when the number of projections is very low.
\textbf{(2)} Methods such as 2D3DNR~\cite{dong20232d} deform a pre-acquired volume of the patient based on input projections. However, the predicted deformation can become  under-constrained when the number of projections is very low.
\textbf{(3)} Methods such as X2Vision~\cite{x2vision} first learn a volume generative model and optimize the parameters of this model to match the projections. However, they are not able to exploit the pre-acquired volume. 
\textbf{(4)} NeRF-based methods~\cite{shen2022nerp} can take the pre-acquired volume as input and optimize on the volume to match the projections---however, they do not exploit any anatomy knowledge besides the pre-acquired volume. 
\textbf{(5)} 
To avoid predicting incorrect deformations when the number of projections gets too low, we propose to guide the deformations using a volume generative model. 
Note that this is different from X2Vision, which relies on a generative model to directly create the predicting volume. Instead, our approach deforms the pre-acquired volume under the guarantee that the resulting volume is anatomically possible.  
}
\label{fig:comparisons}
\end{figure*}

\vincent{
Figure~\ref{fig:comparisons} illustrates existing approaches to decreasing the number of projections when reconstructing a volume. One approach is to train a deep model to regress the volume from a number of projections in a supervised way~\cite{henzler2018single,shen2019patient,ying2019x2ct, jiang2022mfct, tan2022xctnet, Lu_2022, tan2023semi, zhang2023xtransct,  shen2022geometry, zhang2021unsupervised, tian2022liftreg, wang2023tpg}, but such direct inference often results in poor reconstruction. 
}

Instead, X2Vision~\cite{x2vision} learns an anatomical prior and optimizes its parameters to match the projections. Such optimization approaches~(see also \cite{shen2022nerp}) tend to generalize much better. 

In modern medical practice, CT and MRI scans are now widely used for treatment planning and diagnostics.  This provides a volume captured under a different patient pose and  different from the volume to reconstruct by medically-relevant changes such as weight loss or tumor transformation. X2Vision ignores this pre-captured volume but 2D3DNR~\cite{dong20232d} deforms it given several projections.

How can we exploit both sources of information, anatomy knowledge from a generative model and the pre-acquired volume, to improve the reconstruction quality to the point it can actually be used for medical applications? This is the goal of our method. 

Getting the best of these two worlds is however not straightforward to do properly as methods like X2Vision and 2D3DNR work in very different ways.  We start by observing that when 2D3DNR deforms the pre-acquired volume given very few projections, unrealistic deformations are prone to occur. We therefore propose to guide the deformations using a volume generative model.
Note that this is different from X2Vision, which directly optimizes the generative model parameters to match the projections: Our approach allows to deform the pre-acquired volume under the guarantee that the resulting volume is anatomically possible.

To do so, we optimize over the generative model parameters so that the pre-acquired model match well the projections after being mapped to the generated volume.  This is illustrated in Figure~\ref{fig:comparisons}(5). Compared to 2D3DNR for example, we guarantee that the deformed pre-acquired volume is anatomically possible, since it is constrained to be close to a generated (thus anatomically correct) volume. Moreover, we also have the guarantee that the deformed volume match well the projections.  Compared to X2Vision, because our approach predicts the pre-acquired volume after deformation, it captures the patient's unique anatomy or abnormalities accurately. This is by contrast with a generated volume, as done by X2Vision, which often lack critical details.

In our evaluations, we focus on head-and-neck CT scans from cancer patients undergoing regular radiotherapy from cohorts at two different medical centers. The head and neck region is a very challenging part of the human anatomy, maybe the most challenging one, as it is very heterogeneous with complex shapes~(larynx, jaw, teeth, etc.) and its deformations are complex---combination of twist of the neck, articulation of the jaw, and compression and extension due to patient weight and tumor variations.

We compare our method against top-tier techniques, including X2Vision~\cite{x2vision}, 2D3DNR~\cite{dong20232d}, and the NeRF-based approach~\cite{shen2022nerp}. The results demonstrate our method's superiority and ability to capture important medical details. 

Our method demonstrates high-quality rigid registration, indicating a move towards more precise biplanar systems over traditional 3D visualization. Unlike typical 2D/3D registrations focused on bones, our method enables finer adjustments due to our detailed 3D reconstructions. The accuracy of our models, evidenced by high dice coefficients, aligns closely with actual patient anatomy, offering significant improvements for daily treatments and planning. Our approach also reduces irradiation significantly while delivering accurate volume reconstructions to guide procedures, showing potential to transform medical practices.

A portion of the data used in our study is already publicly available. We plan to submit a clinical request to release the longitudinal test dataset as well.

\section{Related Work}

\subsection{3D Reconstruction from few X-Ray Projections}

Many methods have already been proposed to reduce the number of X-ray projections when reconstructing a 3D volume~\cite{henzler2018single,shen2019patient,ying2019x2ct, jiang2022mfct, tan2022xctnet, Lu_2022, tan2023semi, zhang2023xtransct,  shen2022geometry, wang2023tpg, x2vision, shen2022nerp, zha2022naf}. 
A strategy is to train a deep learning model for volume estimation from multiple projections in a supervised manner, as in ~\cite{henzler2018single,shen2019patient,ying2019x2ct, jiang2022mfct, tan2022xctnet, Lu_2022, tan2023semi, zhang2023xtransct,  shen2022geometry, wang2023tpg}. However it requires unchanging sensor calibrations from training to deployment, and this direct volume estimation often yields poor reconstructions.
Most current methods employ a single or multibranch feature embedding from 2D projections~\cite{henzler2018single, shen2019patient}, incorporating a fusion mechanism to predict the 3D volume~\cite{ying2019x2ct, jiang2022mfct, tan2022xctnet, Lu_2022, zhang2023xtransct}, and often integrating adversarial models for enhancement~\cite{ying2019x2ct, jiang2022mfct, wang2023tpg}. To reduce the space of potential solutions, some methods introduce geometric constraints~\cite{shen2022geometry} with refinement of a first estimate with a 3D network like U-Net~\cite{shen2022geometry, Lu_2022}. Recent developments include enriching 2D feature extraction using self-attention mechanisms~\cite{tan2022xctnet} and transformer architectures~\cite{zhang2023xtransct, wang2023tpg}. \cite{tan2023semi} investigates the incorporation of semi-supervised learning within a teacher-student framework, addressing the challenges presented by the limited availability of paired volume-projection data. 
Our method advances beyond these limitations by being fully unsupervised, thereby eliminating the need for paired datasets.

Some methods~\cite{shen2022nerp, x2vision} that iteratively optimize on the reconstructed volume at inference given projections typically perform better. X2Vision~\cite{x2vision} relies on a learned 3D manifold to find a realistic and matching solution, while other rely on NeRF~\cite{shen2022nerp, zha2022naf}. These methods ensure projection consistency at inference and offer superior generalization, thanks to their unsupervised- or non-learning-based  approach. Some of these methods~\cite{shen2022nerp} are even already able to exploit the volume acquired at the beginning of the therapy. However, \cite{shen2022nerp} does this simply by using the pre-acquired volume to initialize the NeRF. When using very few projections, this falls short for effective reconstruction as shown in \cite{x2vision}. Our approach efficiently combines such a pre-acquired volume thanks to a prior on its possible deformation.

\subsection{2D/3D Deformable Registration}

Our approach is also related to 2D/3D deformable registration problems, as we deform the pre-acquired 3D volume by comparing its projections to the captured X-ray projections. 

A common approach to 2D/3D deformable image registration involves solving an optimization problem to find the best transformation parameters that explain the deformation between a volume  and a set of 2D projections, as discussed in \cite{flach2014deformable, prummer2006multi, tian2020fluid, zikic2008deformable}. This process involves evaluating image similarity by comparing the actual CT projections with their corresponding simulated CT projections. Nonetheless, when only a limited number of projections is used, traditional approaches without deep learning struggle.

Some learning-based methods for 2D/3D deformable image registration use feedfoward predictions for faster registration~\cite{foote2019real, li2020non, pei2017non, zhang2021unsupervised}. \cite{zhang2021unsupervised} for instance uses a U-Net structure to predict deformation fields directly from a CT volume and captured X-rays projections, maintaining projection consistency in training. 

This method still struggles with spatial ambiguity in 2D projections due to insufficient 3D spatial information in the training's loss function, essential for learning accurate 3D deformations from high-quality, real 3D pairs, a limitation underscored in \cite{tian2022liftreg}. 
Other works~\cite{pei2017non, li2020non} attempt to overcome this by using prior data to constrain deformations within a realistic range using PCA, yet this does not fully eliminate the spatial ambiguity of 2D measurements. 

The main challenge in 2D/3D deformable registration is resolving spatial ambiguity. Deep learning can be a solution by including accurate 3D data in the training loss function, which may reduce spatial ambiguity and retain generalizability during testing, even without 3D image pairs.
\cite{tian2022liftreg} introduces a method that is able to extract 3D spatial information from the backprojection of projections, further helping the process by incorporating prior knowledge of patient motion via a PCA-reduced space, similar to atlas registration techniques. 
Building on this, \cite{dong20232d} introduced a method we reference by the name ``2D3DNR'' that transitions from 2D biplanar projections to 3D space. This is achieved by estimating a 3D feature map from the projections, followed by a 3D-3D deformation learning process using a U-Net-based model~\cite{cciccek20163d}. 
Our method includes two types of prior: a prior on the predicted volume and a prior on the deformation of the pre-acquired volume. These priors can be learned in a unsupervised way. As our results show, this is key to perform better than the previous state-of-the-art methods.

\begin{figure*}
\centering
\includegraphics[width=1.\linewidth]{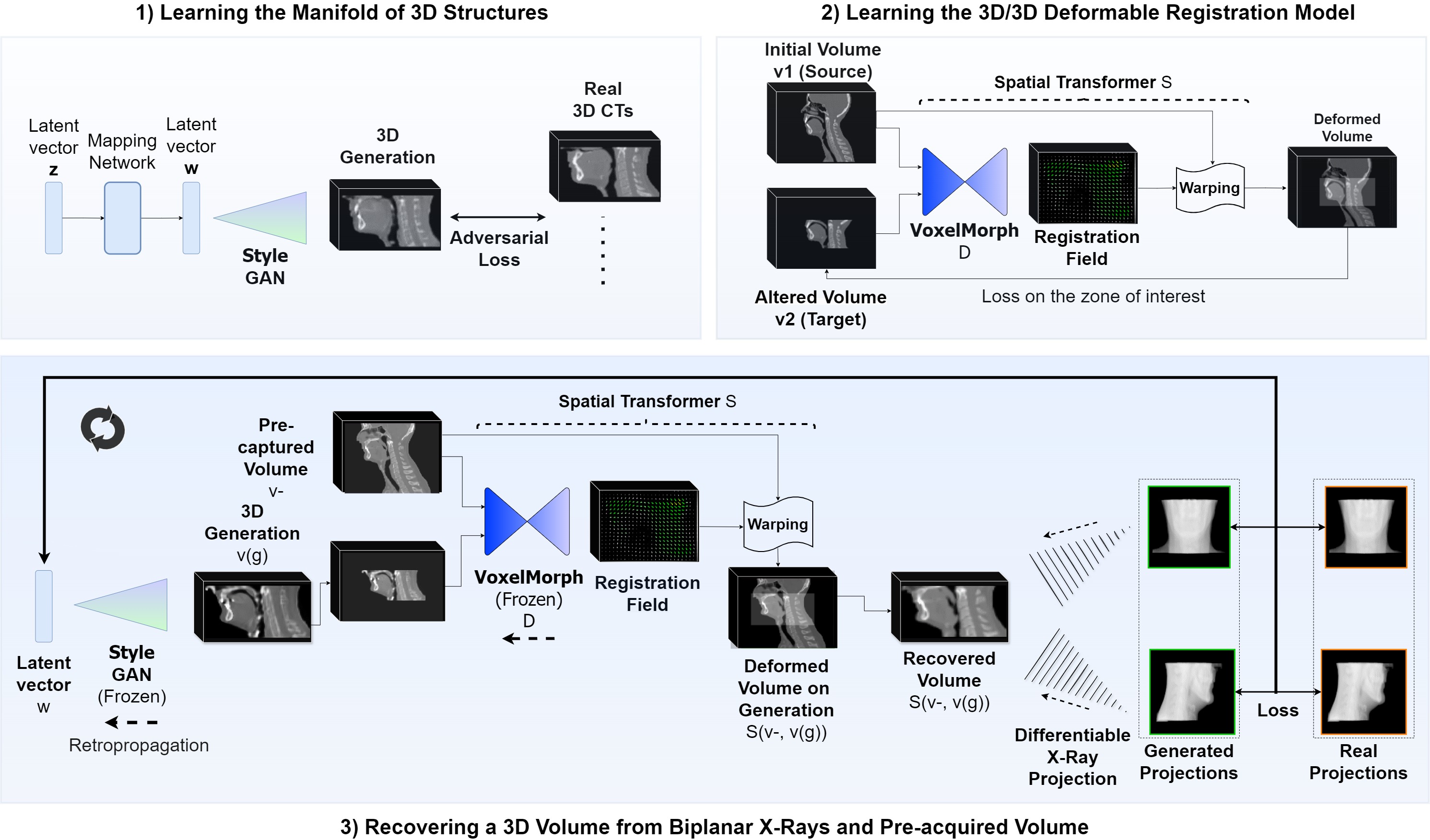}
\caption{\textbf{Our pipeline.} We first train a generative model to generate 3D volumes in a low-dimensional manifold and a 3D/3D deformable registration model between two volumes. Now, given two projections and a pre-captured volume of the patient, we recover a 3D volume corresponding to the two projections by finding the latent vectors that generate the best 3D volume so that the pre-captured volume well deforms on it to match the projections. We iteratively refine the generation and deformations based on the discrepancy between the generated and actual projections. 
}
\label{fig:reconstruction}
\end{figure*}

\section{Method}

In this section, we first formalize our approach, and then describe each of its components.

\subsection{Problem Formulation}\label{problem_formulation}

Given a limited set of projections $\{I_i\}_i$, our objective is to reconstruct the 3D tomographic volume $v$ responsible for these projections. In practice, we consider only two projections as it is medical practice.

Also following medical practice, a previously-captured volume $v^-$ of the patient is available as well. Between $v^-$ and $v$ are both rigid and non-rigid transformations, as well as more complex transformations such as tumor growing or shrinking. We thus seek the transformation of $v^-$ to $v$. 

Finding the deformation is ill-posed in general when the number of projections becomes small. Our main contribution lies in the following formulation that enforces the predicted deformation of $v^-$ to produce an anatomically correct volume:
\begin{equation} \label{eq:one}
\bg^* = \operatorname*{argmin}_{\bg} \sum_i \calL_i(S(v^-, v(\bg)), I_i) + \calR(\bg) \> .
\end{equation}
We briefly describe below each component of this formulation, then describe them in more details in the rest of the section:
\begin{itemize}[noitemsep,topsep=0pt,parsep=0pt,partopsep=0pt,leftmargin=*]
\item $v(.)$ is a generative model of volumes of parameters $\bg$, i.e., $v(\bg)$ is a generated volume. In practice, we use a model similar to the one in X2Vision.
However, X2Vision uses this model to directly predict the final volume $v$.  The key difference in our method is that we use it here to guide the deformation of $v^-$.
\item $S(v^-, v(\bg))$ is a spatial transformer~\cite{jaderberg2015spatial}~\footnote{Not to be confused with Transformers~\cite{vaswani2017attention}.} trained to predict directly the transformation between two volumes. Here, $S(v^-, v(\bg))$ returns volume $v^-$ after deformation to align on the generated volume with the deformation consistent with training data. We use a  spatial transformer very close to the one proposed by VoxelMorph~\cite{balakrishnan2019voxelmorph}.
\item $\calL_i$ is a loss term that compares the projections of deformed volume $S(v^-, v(\bg))$ with the input projections $I_i$.
\end{itemize}
Intuitively, the optimization on $\bg$ generates a volume $v(\bg)$ that guides the deformation of $v^-$ thanks to the first term and exploits prior knowledge on volumes to recover. After optimization, our method returns $S(v^-, v(\bg))$, the pre-acquired volume after deformation.  

Figure~\ref{fig:reconstruction} illustrates our pipeline that implements our approach: 
We first train generative model $v(\bg)$ as well as spatial transformer $S$, both in an unsupervised way. Given two input projections, we then optimize the parameters $\bg$ of the generative model, which gives us deformed volume $S(v^-, v(\bg))$.

By themselves, generative model $v(\bg)$, spatial transformer $S$,  loss term $\calL_i$ are relatively standard. We describe them below for the sake of completeness. We end this section by describing the warm-up step we use to bootstrap optimization of Eq.~\eqref{eq:one}.

\subsection{Generative Model $v(.)$}

Like X2vision~\cite{x2vision}, we learn generative model $v(.)$ using GANs. We decompose parameters $\bg$ into a latent vector $\bw$ and Gaussian noise vectors $\bn = \{\bn_j\}_j$: $\bg = [\bw, \bn]$. Latent vector $\bw \in \calN(\bw|\mu,\sigma)$ is computed from an initial latent vector $\bz \in \calN(0,\mathit{I})$ mapped using a learned network $m$: $\bw = m(\bz)$. $\bw$ controls the global structure of the predicted volumes at different scales by its components $\bw_i$, while the noise vectors $\bn$ allow more fine-grained details. Mean $\mu$ and standard deviation $\sigma$ of the mapped latent space can be computed by mapping over initial latent space $\calN(0,\mathit{I})$ after training. Like \cite{x2vision}, we optimize on the noise vectors $\bn$ as well as they are useful to generate high-resolution details. 

To ensure that the predicted volume remains in the manifold of possible volumes, $\calR(\bg)$ is defined as a sum of regularization terms on $\bw$ and $\bn$:
\begin{equation}
\begin{array}{ll}
  \calR(\bg) &= \calR(\bw, \bn) \\[1mm]
&=  \lambda_w \calL_w(\bw) + \lambda_c \calL_c(\bw) + \lambda_n \calL_n(\bn)  \> .
\end{array}
\end{equation}
Terms $\calL_w(\bw) = - \sum_k{\log\calN(\bw_k|\mu,\sigma)}$ and $\calL_n(\bn) = - \sum_j \log\calN(\bn_j | {\bf 0}, \mathit{I})$ ensure that $\bw$ and $\bn$ remain on their respective distributions learned during training. Term $\calL_c(\bw) = - \sum_{i,j} \log\calM ( \theta_{i,j}| 0, \kappa)$ encourages the $\bw_i$ vectors to be col\-li\-near to keep the generation of coarse-to-fine structures coherent. $\calM(\cdot; \mu, \kappa)$ is the density of the Von Mises distribution of mean $\mu$ and scale  $\kappa$, which we take fixed, and $\theta_{i,j}=\text{arccos}(\frac{\bw_i \cdot \bw_j}{\|\bw_i\|\|\bw_j\|})$ is the angle between vectors $\bw_i$ and $\bw_j$.  The $\lambda_*$ are fixed weights.

\subsection{Spatial Transformer $S$}

We use a spatial transformer $S$ very close to the one introduced by VoxelMorph~\cite{balakrishnan2019voxelmorph}. It can be decomposed into:
\begin{equation}
    S(v_1, v_2) = \calW(v_1, D(v_1, v_2)) \> ,
\end{equation}
where $D(v_1, v_2)$ is a deep network predicting a deformation field from $v_1$ to $v_2$; $\calW(v_1, D(v_1, v_2))$ deforms volume $v_1$ according to the deformation field predicted by $D$. Model $D$ is trained to predict deformation $\calW$ between two volumes $v_1$ and $v_2$ by minimizing
\begin{equation} \label{training_voxelmorph}
\lambda_s \| v_2 -  S(v_1, D(v_1, v_2)) \|^2 + \lambda_D \lVert \nabla D(v_1, v_2))[x] \rVert^2 \>,
\end{equation}
over a training set of corresponding volumes $\{(v_1,v_2)\}$. The second term is a smoothing loss that mitigates sharp local fluctuations and promote smoothness of the predicted field. $\lambda_s$ and $\lambda_D$ are balancing weights that adjust the emphasis between similarity and regularization during training.

Maintaining 1-to-1 mapping in medical image registration is crucial to prevent tearing or overlapping. Our model, inspired by the VoxelMorph approach, predicts a velocity field. By integrating this velocity field over time, we obtain smooth, invertible transformations that naturally avoid singularities. This approach ensures the deformation remains diffeomorphic.

\subsection{Loss Term $\calL_i$}

As in X2Vision, we take term $\calL_i(v, I_i)$ as the weighted sum of the Euclidean distance and the perceptual loss~\cite{johnson2016perceptual}:
\begin{equation}
\calL(v, I_i) = \lambda_2 \big\|A_i \circ v - I_i \big\|_2 
    + \lambda_p \calL_p(A_i \circ v, I_i) \> ,
\end{equation}
as we observed that this combination results in the best results.
$A_i$ denotes an operator that projects volume $v$ under view $i$---we detail it in the supplementary material.

\subsection{Warm-Up}

Before optimizing Eq.~\eqref{eq:one} we first retrieve an initial volume estimate $v(\bg)$ by performing several gradient descent steps of objective
\begin{equation} \label{eq:initialization}
\sum_i \calL_i(v(\bg), I_i) + \calR(\bg) \> ,
\end{equation}
starting from random initialization for $\bg$. We use 10 iterations in practice. This provides a better initialization for $\bg$ before optimizing Eq.~\eqref{eq:one} and speeds up convergence.


\section{Experiments}

We evaluate our method for our target application, head-and-neck cancer radiotherapy. Head-and-neck imagery exhibits many fine details and complex deformations and is representative of many of the different challenges of volume recovery. In this section, we introduce our dataset, models for learning key priors, and present both quantitative and qualitative comparisons with state-of-the-art methods. We also include an ablation study to evaluate the contribution of our priors and an analysis of our run-time efficiency of our method.

\newlength\results  \setlength\results{14.5mm} 
\newlength\resultsAdj  \setlength\resultsAdj{14mm} 

\begin{figure*}
\includegraphics[clip, width=1\textwidth]{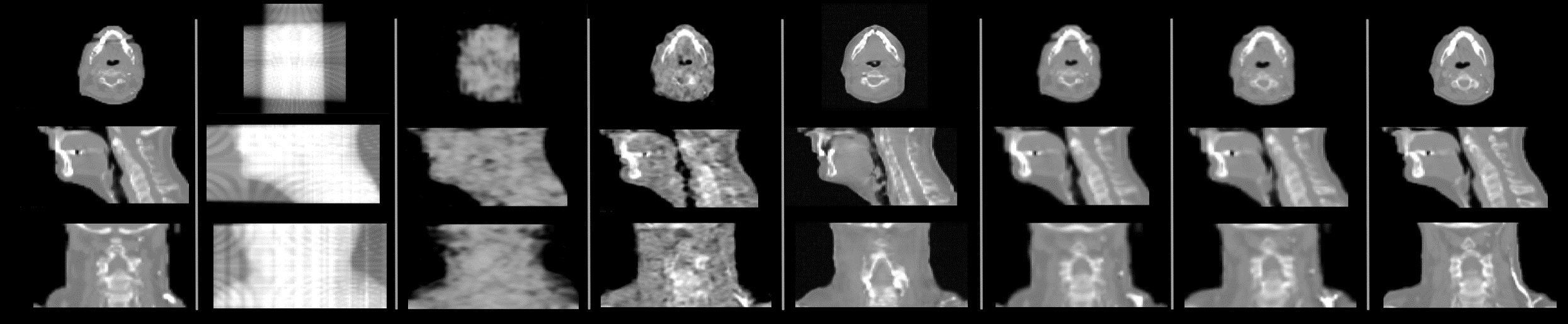}
\includegraphics[clip, width=1\textwidth]{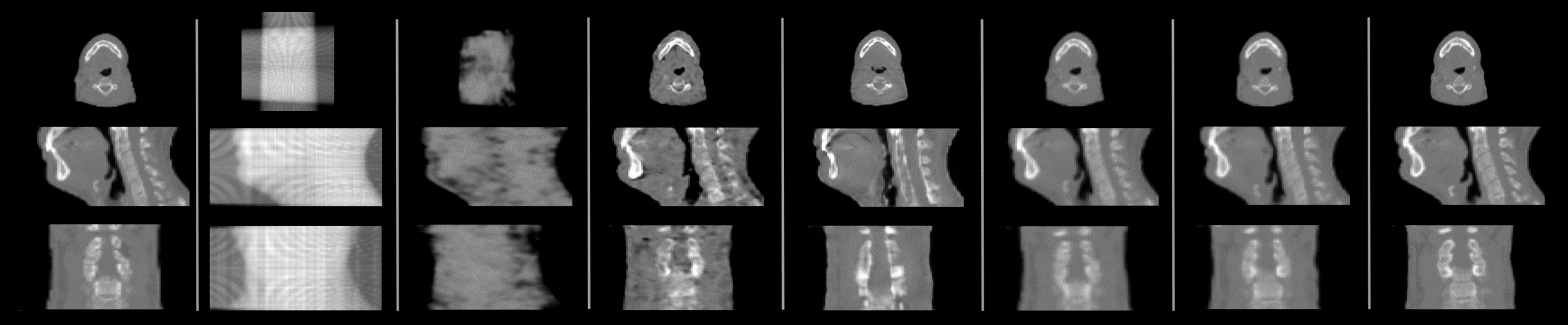}
\includegraphics[clip, width=1\textwidth]{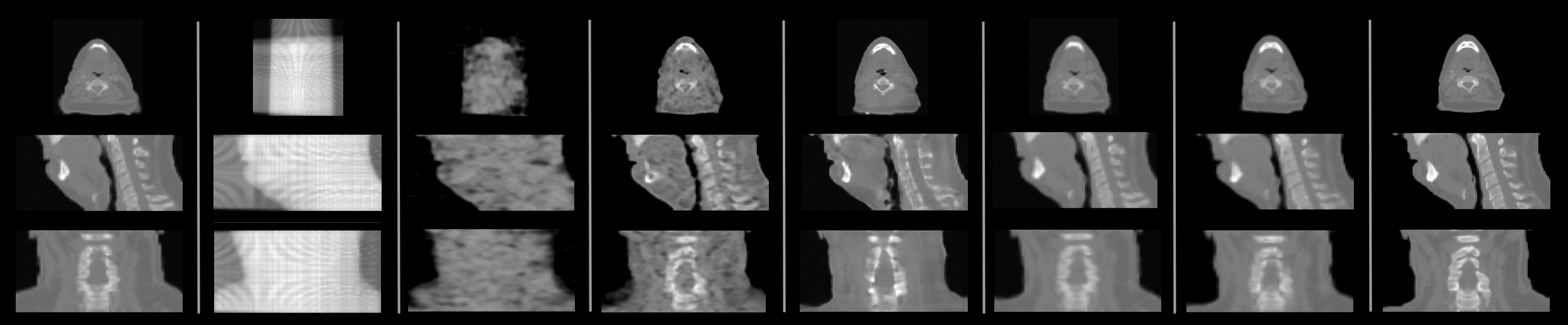}
\scriptsize
\begin{tabular}{@{}C{\results}C{\results}C{\results}C{\resultsAdj}C{\results}C{\results}C{\results}C{\results}
@{}}
\\[-0.1cm]
pre-captured CT &  back\-projec\-tion & NeRP~\cite{shen2022nerp} w/o & NeRP~\cite{shen2022nerp} w/ &  X2Vision~\cite{x2vision} & 2D3DNR~\cite{dong20232d} & Ours  & Ground Truth \\ 
\end{tabular}
\caption{\label{fig:rec} 
\textbf{Visual analysis of recovered volumes from two projections by previous methods and our approach.} In the absence of a pre-captured CT volume, NeRP struggles due to lack of  constraints. When exploiting the pre-captured CT volume, NeRP still tends to introduce artifacts in an attempt to align with the projections and alters the anatomy without ensuring anatomical accuracy. In contrast to X2Vision, our method predicts a reconstruction that captures patient-specific details and nuances. 2D3DNR results in deformations that do not adequately match the anatomy.
}
\label{visuals}
\end{figure*}

\subsection{Datasets and Metrics}

\paragraph{Volume Generator Learning.} 

We trained our GAN model for $v(\bg)$ on a large dataset of 3500 CTs of patients with head-and-neck cancer, more exactly 2297 patients from the publicly available The Cancer Imaging Archive~(TCIA)~\cite{HNSCC,radiomics,head_neck,qin,acrin,tcga} and 1203 from private internal data, after obtaining ethical approbation. We split this data into 3000 cases for training, 250 for validation, and 250 for testing. 

We targeted CT scans of the head-and-neck region with a resolution of $80 \times 96 \times 112$. Using an automatic segmentation of the mouth with a pre-trained U-Net~\cite{ronneberger2015u}, we centered these scans around it. We applied min-max normalization and post-clipping values between -1024 and 2000 Hounsfield Units~(HU) to the CT scans as pre-processing. Our goal is to reconstruct the head-and-neck area using only two projections and the pre-acquired volume.

\paragraph{Longitudinal Radiotherapy Data.}
In radiotherapy, a ``planning CT'' capture is used to design the therapy plan, while daily 3D Cone-Beam CTs~(CBCTs) are captured during treatment to ensure the patient's positioning matches this initial CT. 
With patient consent, we compiled planning CT scans and subsequent CBCT scans from 242 patients across two medical centers, one contributing 177 and the other 65 cases. These datasets, distinct in protocols and scanning equipments, offered a diverse basis for assessing our methods in varied clinical settings.

As depicted in Figure~\ref{fig:rec} and the supplementary material, notable differences emerge between the initial CT scans and subsequent CBCT scans, because of both treatment-induced alterations and patient pose variations. By aligning planning CTs with CBCTs using MRF minimization~\cite{glocker2008dense}, we derived 3D volumes as CTs. We used some of these volumes for training our 3D/3D deformable registration model and the rest for evaluating the predicted reconstructions by the different methods.

\paragraph{3D/3D Deformable Registration Training.}
More precisely, to train our 3D/3D deformable registration model, we randomly selected 146 patients for training, 16 for validation, and 10 for testing. We paired each initial CT with any subsequent CT from the same patient to obtain a large training set.

\paragraph{Volume Recovery.}
The second part of volumes, used for evaluation, includes 70 patients showcasing the most marked longitudinal alterations. These were selected by comparing their CBCTs with planning CTs. We paired the planning CT with each patient's final CT---which underscores the utmost discrepancies. We used the planning CT as pre-captured volume. Projections were derived from the last 3D volumes, focusing on the reconstruction area, using our projector detailed in previous section.

\paragraph{Metrics.}\label{metrics}
We assessed the reconstruction performance using two quantitative metrics: PSNR, which quantifies reconstruction error, and SSIM, which gauges the perceptual quality of the images. 

We also evaluated the accuracy of the deformation between the pre-acquired volume and the recovered volume for the two methods that estimate this deformation: 2D3DNR and ours. To this end, we consider the Dice score for the mouth and the larynx, two structures that are likely to deform significantly. To compute it, we segmented these structures on the pre-captured volumes and the recovered volumes using a trained U-Net model using about 1000 head-and-neck CTs.

Additionally, we compared the 3D rigid registration differences between the initial full CT scans and our reconstructions against the ground truth, including variations in rotation angles and translations across all axes. This comprehensive analysis helps to underline the precision of our method in capturing and reconstructing the nuanced deformations of critical anatomical features.

More implementation details are provided in the supplementary material.


\begin{table}
\centering
\caption{\textbf{Metrics on volumes  from two projections by previous methods and our approach.}
Standard deviations are provided in parentheses. (w/) and (w/o) stand for the use or not of the pre-captured volume respectively.
}
\label{tab:metrics}
\addtolength{\tabcolsep}{+1pt}
\scalebox{0.85}{
\begin{tabular}{@{}c@{$\quad$}c@{}}
\begin{tabular}{@{}lcc@{}}
\toprule
\multicolumn{1}{c}{Method} & PSNR~(dB) $\uparrow$ &  SSIM $\uparrow$ \\ 
\midrule
Backprojection & $10.29~(\pm 0.5)$ & $0.23~(\pm 0.01)$ \\[1.1mm]
NeRP (w/o)~\cite{shen2022nerp}  & $19.81~(\pm 1.7)$ & $0.21~(\pm 0.03)$ \\[1.mm]
NeRP (w/)~\cite{shen2022nerp} & $25.32~(\pm 1.6)$ & $0.34~(\pm 0.02)$ \\[1.mm]
X2Vision~\cite{x2vision} & $27.80~(\pm 1.4)$ & $0.89~(\pm 0.03)$ \\[1.mm]
2D3DNR~\cite{dong20232d} & $29.07~(\pm 1.6)$ & $0.92~(\pm 0.02)$ \\
\midrule
Ours & $\mathbf{33.23~(\pm 0.62)}$ & $\mathbf{0.96~(\pm 0.01)}$ \\
\bottomrule
\end{tabular} 
&
\begin{tabular}{@{}lcc@{}}
\toprule
\multicolumn{1}{c}{\multirow{2}{*}{Method}} & \multicolumn{2}{c}{Dice $\uparrow$} \\
\cline{2-3}\multicolumn{1}{c}{} & Mouth & Larynx\\ 
\midrule
2D3DNR~\cite{dong20232d} & $0.91~(\pm 0.03)$ & $0.80~(\pm0.07)$ \\[1.mm]
Ours & $\mathbf{0.95~(\pm 0.01)}$ & $\mathbf{0.91~(\pm0.02)}$ \\
\bottomrule
\bottomrule
\multicolumn{1}{c}{\multirow{2}{*}{Method}} & \multicolumn{2}{c}{Rigid Registration Error (6 DoF)} \\
\cline{2-3}
\multicolumn{1}{c}{} & Rotation~(°) $\downarrow$ & Translation~(mm) $\downarrow$\\ 
\midrule
2D3DNR~\cite{dong20232d} & $0.52~(\pm 0.29)$ & $0.88~(\pm0.45)$ \\[1.mm]
X2Vision~\cite{x2vision} & $0.45~(\pm 0.31)$ & $0.50~(\pm0.26)$ \\[1.mm]
Ours & $\mathbf{0.16~(\pm 0.15)}$ & $\mathbf{0.20~(\pm0.07)}$ \\
\bottomrule
\end{tabular}
\end{tabular}
}
\end{table}

\subsection{Results and Analysis}
\label{results}

Table~\ref{tab:metrics} reports the quantitative results. We detail below the methods we compare to and discuss their results after they were retrained on our data. Figure~\ref{fig:rec} compares visually our reconstruction to these methods on several examples.  Additional results and reconstructions are provided in the supplementary material, but we summarize below our visual analysis of the results.

The \textbf{backprojection method} is a very simple baseline inspired from \cite{FDK}. It estimates the value of each voxel as the average of the values at the projected voxel locations in the input X-ray projections. When enough input projections are available, this method can provide satisfying results. However, it fails when only two projections are used.

The \textbf{NeRP} method~\cite{shen2022nerp} optimizes the 3D volume to match the projections. It also struggles when very few projections are given since they lack prior anatomical knowledge. Even when conditioned on the pre-captured volume, it is often not able to eliminate the many artefacts.

We also considered the recent \textbf{X2Vision}~\cite{x2vision} method to highlight the advantages of exploiting the pre-captured volume as we do---which X2Vision does not. It provides a reasonable reconstruction but still misses important details.

\textbf{2D3DNR}~\cite{dong20232d} predicts in a feedforward way the deformation between the pre-captured 3D volume and the new one given the precaptured volume itself and the available projections. Since the original code was unavailable, we used the same VoxelMorph backbone as ours to reimplement the 3D/3D registration method. Further details can be found in the supplementary material. The  volumes predicted by 2D3DNR do not reproject well on the input projections in general. Because it is a feedforward method, it also tends to generalize poorly. Our method recovers better the deformation of the tissues.

Like X2Vision and NeRP, our method optimizes on the volume during inference for consistency with the input projections, which helps generalization. It also introduces a prior on the anatomical volume thanks to its GANs, in a way related to X2Vision. Our method has however an original way to exploit the pre-acquired volume by controling its deformations. This contrast with 2D3DNR, which takes this volume as input to a feedforward process, and with NeRP, which uses this volume only as conditioning. Our approach appears to be more powerful as it yields the best results.

Although X2Vision provides a capable reconstruction, it lacks in detailing critical aspects. Our method emphasizes integrating patient-specific details to surpass the constraints of the generative model manifold, which might not capture the patient's unique anatomy or abnormalities accurately. By leveraging the pre-acquired volume, our method obtains a more accurate depiction of the patient's real anatomy rather than depending on a generalized learned manifold. This focus on patient specificity is crucial for achieving detailed and lifelike anatomy reconstructions.

In stark contrast to 2D3DNR, our approach adopts an optimization strategy to inform the deformation prior, significantly enhanced by the capabilities of the generative model. This model lays down a realistic 3D scaffold for the optimization process, ensuring the deformations are not just plausible but supported by the anatomical description of the generative model. This leads to reconstructions that are markedly more precise, showcasing a significant leap forward in the fidelity of anatomical reconstruction techniques.

\subsection{Validation for Medical Applications}

As shown on the right of Table~\ref{tab:metrics}, our method demonstrates superior deformation accuracy.
Moreover, we attain impressive rigid registration with an average precision way below 1mm. This level of detail and accuracy underscores the effectiveness of our approach in capturing the complex nuances of patient anatomy. Such high-quality volume reconstruction from only biplanar projections opens the way for accurate 3D rigid registration and daily assessment of the evolution of the patient, without needing 3D acquisition.  

\begin{table}
\centering
\caption{
\textbf{Ablation Study.} This table shows the contribution of our two priors. 
'deformation of the pre-acquired volume without any prior' only uses the pre-acquired volume and no prior.
'generative model only' does not use the pre-acquired volume. 
'generative model followed by 1 deformation' uses the pre-acquired volume only after the independent reconstruction phase conducted by the generative model.
'deformation of the pre-acquired volume only' only uses the pre-acquired volume and the prior on its deformations. 
More details about these experiments are given in Section~\ref{sec:ablation}. Standard deviations are provided in parentheses.
}
\addtolength{\tabcolsep}{+1pt}
\scalebox{0.9}{
\begin{tabular}{@{}lcc@{}}
\toprule
\multicolumn{1}{c}{Method} & PSNR~(dB) $\uparrow$ &  SSIM $\uparrow$ \\ 
\midrule
deformation of the &&\\[-0.2mm]
pre-acquired volume & $27.04~(\pm 1.9)$ & $0.88~(\pm 0.03)$ \\[-0.2mm]
without any prior  & & \\[1.2mm]
generative model only & $27.80~(\pm 1.4)$ & $0.88~(\pm 0.03)$ \\[1.2mm]
generative model & \multirow{2}{*}{$29.24~(\pm 1.8)$} & \multirow{2}{*}{$0.92~(\pm 0.02)$} \\[-0.2mm]
followed by 1 deformation & & \\[1.2mm]
deformation of the &\multirow{2}{*}{$30.75~(\pm 1.19)$} & \multirow{2}{*}{$0.93~(\pm 0.01)$}\\[-0.2mm]
pre-acquired volume only && \\[1.2mm]
full method & $\mathbf{33.23~(\pm 0.62)}$ & $\mathbf{0.96~(\pm 0.01)}$ \\

\bottomrule
\end{tabular}
}
\label{tab:ablation_study}
\end{table}

\newlength\lengthmini \setlength\lengthmini{4mm} 
\newlength\lengthaa \setlength\lengthaa{6.5mm} 
\newlength\lengthbb \setlength\lengthbb{4.5mm} 
\newlength\lengthcc \setlength\lengthcc{5.5mm} 
\newlength\lengthdd \setlength\lengthdd{6mm} 
\newlength\lengthee \setlength\lengthee{8.4mm} 
\newlength\lengthff \setlength\lengthff{13.1mm} 
\newlength\lengthgg \setlength\lengthgg{10mm} 
\newlength\lengthhh \setlength\lengthhh{10mm} 

\newlength\ablation  \setlength\ablation{16.4mm} 

\begin{figure*}
\includegraphics[clip, width=1\textwidth]{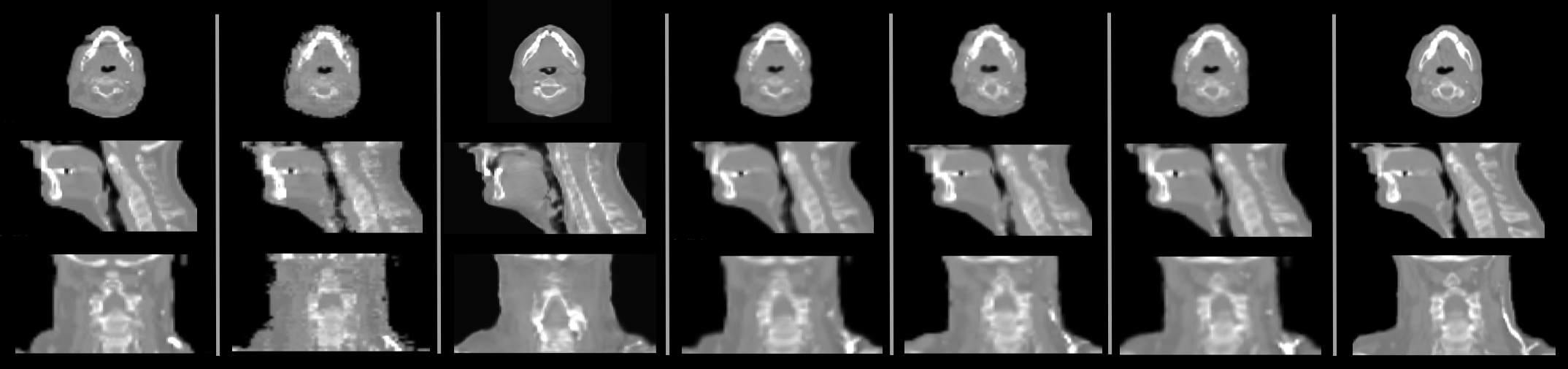}
\scriptsize
\begin{tabular}{@{}C{\ablation}C{\ablation}C{\ablation}C{\ablation}C{\ablation}C{\ablation}C{\ablation}@{}}
\\[-0.1cm]                
pre-captured CT &  deformation of the pre-acquired volume without any prior & generative model only & generative model followed by 1 deformation  &  deformation of the pre-acquired volume only & full method  & Ground Truth \\ 
\end{tabular}
\caption{\label{fig:ablation} 
\textbf{Visual Analysis of the Ablation Study.} Deforming the pre-acquired volume without any prior results in erratic and anatomically inconsistent changes. Reconstruction solely with the generative model may overlook details and lead to mismatches. Deforming the pre-acquired volume on it introduces patient-specific features but may retain initial misalignments.  While introducing prior on deformation aids guiding the direction process, it leads to unnatural distortions of body contour and bone structures. 
Our method, by leveraging both anatomical and deformation priors, yields more realistic and anatomically preserving results.
}
\label{visuals}
\end{figure*}

\subsection{Ablation Study}
\label{sec:ablation}

Table~\ref{tab:ablation_study} presents an ablation study highlighting the benefits of our loss function in Eq.~\eqref{eq:one} by comparing it to different possible variants. Figure~\ref{fig:ablation} presents a visual comparison of the results obtained with these variants.
We considered four variants. The reader should compare the loss functions for these variants to the loss function we introduced in Eq.~\eqref{eq:one}:
\begin{itemize}[noitemsep,topsep=0pt,parsep=0pt,partopsep=0pt,leftmargin=*]
\item `deformation of the pre-acquired volume without any prior': This variant returns volume $\calW(v^-, \phi^*)$ with:
\begin{equation}
\phi^* = \operatorname*{argmin}_{\phi} \sum_i \calL_i(\calW(v^-, \phi), I_i) \> ,
\end{equation}
where $\calW(v^-, \phi)$ applies a deformation field $\phi$ to the pre-acquired volume $v^-$. This approach is related to 2D3DNR as 2D3DNR also deforms the pre-acquired volume. The difference is that 2D3DNR predicts the deformation in a feedforward fashion while this approach retrieves the deformation parameters iteratively.
The retrieved deformations tend to be erratic, blending structures and leading to artifacts that compromise anatomical accuracy. \\[0.1cm]
\item `generative model only': This variant returns $v(\bg^*)$ with
\begin{equation} \label{eq:generative_only}
\bg^* = \operatorname*{argmin}_{\bg} \sum_i \calL_i(v(\bg), I_i) + \calR(\bg) \> .
\end{equation}
It uses only the generative model to predict the volume  and corresponds to the X2Vision method.  \\[0.1cm]

\item `generative model followed by 1 deformation': This variant returns volume $S(v^-,v(\bg^*))$ with $\bg^*$ retrieved by optimizing Eq.~\eqref{eq:generative_only}. This approach deforms the pre-acquired volume to fit the generative model's reconstruction, introducing patient-specific details but potentially retaining initial mismatches. This shows the advantage of combining volume $v^-$ and the generative model $v(\bg)$ during optimization.  \\[0.1cm]
\item `deformation of the pre-acquired volume only': This variant returns volume $S(v^-,\bv^*)$ with:
\begin{equation}
\bv^* = \operatorname*{argmin}_{\bv} \sum_i \calL_i(S(v^-, \bv), I_i) \> ,
\end{equation}
where $\bv$ is a volume represented by a voxel grid, with each voxel encompassing an intensity to optimize. 
This approach uses only the pre-acquired volume and the spatial transformer, but not the generative model. This results in local deformations that are not anatomically realistic, such as bone extensions or body contour distortions, stemming from its lack of anatomical prior.
\end{itemize}

\noindent Further details are provided in the supplementary material. 
The quantitative results clearly show that our loss function exploits both  priors well.

\subsection{Inference Time}
Due to lack of space, a comparison of inference times for the different methods is given in the supplementary material.  Our method recovers high-quality volumes in only 1 minute. While some other methods are faster, the trade-off fidelity/runtime is well acceptable as clinical CBCT acquisition and FDK reconstruction~\cite{FDK} currently requires more than 2 minutes.

\section{Conclusion}

Our studies show that merging two key priors into one optimization problem significantly outperforms existing techniques. Using patient-specific data and anatomical constraints, we achieve unmatched accuracy in anatomical reconstructions, avoiding the need for intensive 3D scans. This method promises improved patient care with daily adjustable treatments like adaptive radiotherapy, enhancing precision and outcomes while reducing treatment times and radiation exposure. Such advancements have the potential to revolutionize healthcare by enabling safer and more personalized treatment approaches.

\newpage


{\small
\bibliographystyle{splncs04}
\bibliography{biblio}

\begin{thebibliography}{10}
\providecommand{\url}[1]{\texttt{#1}}
\providecommand{\urlprefix}{URL }
\providecommand{\doi}[1]{https://doi.org/#1}

\bibitem{balakrishnan2019voxelmorph}
Balakrishnan, G., Zhao, A., Sabuncu, M.R., Guttag, J., Dalca, A.V.: {VoxelMorph: A Learning Framework for Deformable Medical Image Registration}. IEEE Transactions on Medical Imaging  \textbf{38}(8) (2019)

\bibitem{qin}
Beichel, R.R., Ulrich, E.J., Bauer, C., Wahle, A., Brown, B., Chang, T., Plichta, K.A., Smith, B.J., Sunderland, J.J., Braun, T., Fedorov, A., Onken, D.C.M., Magnotta, V.A., Menda, Y., Riesmeier, J., Pieper, S., Kikinis, R., Graham, M.M., Casavant, T., Sonka, M., Buatti, J.M.: {Data from QIN-HEADNECK}  (2015)

\bibitem{x2vision}
Cafaro, A., Spinat, Q., Leroy, A., Maury, P., Munoz, A., Beldjoudi, G., Robert, C., Deutsch, E., Gr\'egoire, V., Lepetit, V., Paragios, N.: {X2Vision: 3D CT Reconstruction from Biplanar X-Rays with Deep Structure Prior}. In: International Conference on Medical Image Computing and Computer-Assisted Intervention (2023)

\bibitem{cciccek20163d}
{\c{C}}I{\c{c}}ek, {\"O}., Abdulkadir, A., Lienkamp, S.S., Brox, T., Ronneberger, O.: {3D U-Net: Learning Dense Volumetric Segmentation from Sparse Annotation}. In: Conference on Medical Image Computing and Computer Assisted Intervention (2016)

\bibitem{dong20232d}
Dong, G., Dai, J., Li, N., Zhang, C., He, W., Liu, L., Chan, Y., Li, Y., Xie, Y., Liang, X.: {2D/3D Non-Rigid Image Registration via Two Orthogonal X-Ray Projection Images for Lung Tumor Tracking}. Bioengineering  \textbf{10}(2) (2023)

\bibitem{FDK}
Feldkamp, L.A., Davis, L.C., Kress, J.W.: {Practical Cone-Beam Algorithm}. Science  (1984)

\bibitem{flach2014deformable}
Flach, B., Brehm, M., Sawall, S., Kachelrie{\ss}, M.: {Deformable 3D--2D Registration for CT and Its Application to Low Dose Tomographic Fluoroscopy}. Physics in Medicine \& Biology  \textbf{59}(24) (2014)

\bibitem{foote2019real}
Foote, M.D., Zimmerman, B.E., Sawant, A., Joshi, S.C.: {Real-Time 2D-3D Deformable Registration with Deep Learning and Application to Lung Radiotherapy Targeting}. In: Information Processing in Medical Imaging (2019)

\bibitem{glocker2008dense}
Glocker, B., Komodakis, N., Tziritas, G., Navab, N., Paragios, N.: {Dense Image Registration through MRFs and Efficient Linear Programming}. In: Medical Image Analysis (2008)

\bibitem{HNSCC}
Grossberg, A., Elhalawani, H., Mohamed, A., Mulder, S., Williams, B., White, A.L., Zafereo, J., Gunn, G.B., Frank, S.J., Rosenthal, D.I., Garden, A.S., Fuller, C.D.: {Anderson Cancer Center Head and Neck Quantitative Imaging Working Group. HNSCC}  (2020)

\bibitem{henzler2018single}
Henzler, P., Rasche, V., Ropinski, T., Ritschel, T.: {Single-Image Tomography: 3D Volumes from 2D Cranial X-Rays}. In: Computer Graphics Forum (2018)

\bibitem{jaderberg2015spatial}
Jaderberg, M., Simonyan, K., Zisserman, A., Kavukcuoglu, K.: {Spatial Transformer Networks}. In: Advances in Neural Information Processing Systems (2015)

\bibitem{jiang2022mfct}
Jiang, Y.: {MFCT-GAN: Multi-Information Network to Reconstruct CT Volumes for Security Screening}. Journal of Intelligent Manufacturing and Special Equipment  (2022)

\bibitem{johnson2016perceptual}
Johnson, J., Alahi, A., Fei-Fei, L.: {Perceptual Losses for Real-Time Style Transfer and Super-Resolution}. In: European Conference on Computer Vision (2016)

\bibitem{acrin}
Kinahan, P., Muzi, M., Bialecki, B., Coombs, L.: {Data from the ACRIN 6685 Trial HNSCC-FDG-PET/CT} (2020)

\bibitem{radiomics}
Kwan, J.Y.Y., Su, J., Huang, S.H., Ghoraie, L.S., Xu, W., Hope, A.J., Aerts, H.J., Waldron, J.N., Haibe-Kains, B., O'sullivan, B., Bratman, S.V., Liu, F.F.: {Data from Radiomic Biomarkers to Refine Risk Models for Distant Metastasis in Oropharyngeal Carcinoma}  (2019)

\bibitem{li2020non}
Li, P., Pei, Y., Guo, Y., Ma, G., Xu, T., Zha, H.: {Non-Rigid 2D-3D Registration Using Convolutional Autoencoders}. In: International Symposium on Biomedical Imaging (2020)

\bibitem{Lu_2022}
Lu, S., Li, S., Wang, Y., Zhang, L., Hu, Y., Li, B.: {Prior Information-Based High-Resolution Tomography Image Reconstruction from a Single Digitally Reconstructed Radiograph}. Physics in Medicine \& Biology  \textbf{67}(8) (Apr 2022)

\bibitem{pei2017non}
Pei, Y., Zhang, Y., Qin, H., Ma, G., Guo, Y., Xu, T., Zha, H.: {Non-Rigid Craniofacial 2D-3D Registration using CNN-Based Regression}. In: Deep Learning in Medical Image Analysis and Multimodal Learning for Clinical Decision Support: Third International Workshop, DLMIA 2017, and 7th International Workshop, ML-CDS 2017, Held in Conjunction with MICCAI (2017)

\bibitem{prummer2006multi}
Pr\"ummer, M., Hornegger, J., Pfister, M., D\"orfler, A.: {Multi-Modal 2D-3D Non-Rigid Registration}. In: Medical Imaging 2006: Image Processing (2006)

\bibitem{ronneberger2015u}
Ronneberger, O., Fischer, P., Brox, T.: {U-Net: Convolutional Networks for Biomedical Image Segmentation}. In: Conference on Medical Image Computing and Computer Assisted Intervention (2015)

\bibitem{shen2022nerp}
Shen, L., Pauly, J., Xing, L.: {NeRP: Implicit Neural Representation Learning with Prior Embedding for Sparsely Sampled Image Reconstruction}. IEEE Transactions on Neural Networks  (2022)

\bibitem{shen2022geometry}
Shen, L., Zhao, W., Capaldi, D., Pauly, J., Xing, L.: {A Geometry-Informed Deep Learning Framework for Ultra-Sparse 3D Tomographic Image Reconstruction}. Computers in Biology and Medicine  (2022)

\bibitem{shen2019patient}
Shen, L., Zhao, W., Xing, L.: {Patient-Specific Reconstruction of Volumetric Computed Tomography Images from a Single Projection View via Deep Learning}. Nature  \textbf{3}(11) (2019)

\bibitem{tan2022xctnet}
Tan, Z., Li, J., Tao, H., Li, S., Hu, Y.: {XctNet: Reconstruction Network of Volumetric Images from a Single X-Ray Image}. Computerized Medical Imaging and Graphics  \textbf{98} (2022)

\bibitem{tan2023semi}
Tan, Z., Li, S., Hu, Y., Tao, H., Zhang, L.: {Semi-XctNet: Volumetric Images Reconstruction Network from a Single Projection Image via Semi-Supervised Learning}. Computers in Biology and Medicine  \textbf{155} (2023)

\bibitem{tian2022liftreg}
Tian, L., Lee, Y.Z., San Jos\'e~Est\'epar, R., Niethammer, M.: {LiftReg: Limited Angle 2D/3D Deformable Registration}. In: International Conference on Medical Image Computing and Computer-Assisted Intervention (2022)

\bibitem{tian2020fluid}
Tian, L., Puett, C., Liu, P., Shen, Z., Aylward, S.R., Lee, Y.Z., Niethammer, M.: {Fluid Registration Between Lung CT and Stationary Chest Tomosynthesis Images}. In: Conference on Medical Image Computing and Computer Assisted Intervention (2020)

\bibitem{head_neck}
Valli\`eres, M., Perrin, E.K.R.a.L.J., Liem, X., Furstoss, C., Khaouam, N., Nguyen-Tan, P.F., Wang, C.S., Sultanem, K.: {Data from Head-Neck-PET-CT}  (2020)

\bibitem{vaswani2017attention}
Vaswani, A., Shazeer, N., Parmar, N., Uszkoreit, J., Jones, L., Gomez, A.N., Kaiser, {\L}., Polosukhin, I.: {Attention Is All You Need}. In: Advances in Neural Information Processing Systems (2017)

\bibitem{wang2023tpg}
Wang, Y., Xia, Q.: {TPG-rayGAN: CT Reconstruction Based on Transformer and Generative Adversarial Networks}. In: Third International Conference on Intelligent Computing and Human-Computer Interaction (ICHCI 2022) (2023)

\bibitem{ying2019x2ct}
Ying, X., Guo, H., Ma, K., Wu, J., Weng, Z., Zheng, Y.: {X2CT-GAN: Reconstructing CT from Biplanar X-Rays with Generative Adversarial Networks}. In: Conference on Computer Vision and Pattern Recognition (2019)

\bibitem{zha2022naf}
Zha, R., Zhang, Y., Li, H.: {{NAF}: Neural Attenuation Fields for Sparse-View {CBCT} Reconstruction}. In: Conference on Medical Image Computing and Computer Assisted Intervention (2022)

\bibitem{zhang2023xtransct}
Zhang, C., Dai, J., Wang, T., Liu, X., Chan, Y., Liu, L., He, W., Xie, Y., Liang, X.: {XTransCT: Ultra-Fast Volumetric CT Reconstruction Using Two Orthogonal X-Ray Projections via a Transformer Network}. In: arXiv Preprint (2023)

\bibitem{zhang2021unsupervised}
Zhang, Y.: {An Unsupervised 2D--3D Deformable Registration Network (2D3D-RegNet) for Cone-Beam CT Estimation}. Physics in Medicine \& Biology  \textbf{66}(7) (2021)

\bibitem{zikic2008deformable}
Zikic, D., Groher, M., Khamene, A., Navab, N.: {Deformable Registration of 3D Vessel Structures to a Single Projection Image}. In: Medical imaging 2008: image processing (2008)

\bibitem{tcga}
Zuley, M.L., Jarosz, R., Kirk, S., Lee, Y., Colen, R., Garcia, K., Delbeke, D., Pham, M., Nagy, P., Sevinc, G., Goldsmith, M., Khan, S., Net, J.M., Lucchesi, F.R., Aredes, N.D.: {The Cancer Genome Atlas Head-Neck Squamous Cell Carcinoma Collection~(TCGA-HNSC)}  (2015)

\end{thebibliography}
}

\end{document}